\documentclass[authordate]{miri-tech-article}

\bibliography{MIRIBibliography.bib,MIRIBibliography-temp.bib}

\usepackage{amsmath,amssymb,amsthm,amsopn,amsfonts}
\usepackage[ruled,vlined,noend]{algorithm2e}
\usepackage{tikz}
\usepackage[colorinlistoftodos,textsize=small]{todonotes}
\usetikzlibrary{shapes,arrows}
\usetikzlibrary{positioning,intersections}

\title{Toward Idealized Decision Theory}

\author{Nate Soares \and Benja Fallenstein \\
Machine Intelligence Research Institute\\
\{nate,benja\}@intelligence.org\\}

\newcommand{\set}[1]{\ensuremath{\{\,#1\,\}}}

\newcommand{\Quote}[1]{\ensuremath{\ulcorner#1\urcorner}}
\newcommand{\PA}{\ensuremath{\mathcal{P}\!\mathcal{A}}\xspace}

\newcommand{\Figure}[1]{Figure~\ref{fig:#1}}

\newcommand{\Section}[1]{Section~\ref{sec:#1}}
\newcommand{\Table}[1]{Table~\ref{tab:#1}}

\newcommand{\Pay}{\ensuremath{\mathrm{Pay}}\xspace}
\newcommand{\Refuse}{\ensuremath{\mathrm{Refuse}}\xspace}

\newcommand{\Algo}{\ensuremath{\texttt{A()}}}
\newcommand{\UDT}{\ensuremath{\mathrm{UDT}}}
\newcommand{\Env}{\ensuremath{\texttt{E()}}}
\newcommand{\Hi}{\ensuremath{\mathtt{High}}}
\newcommand{\Med}{\ensuremath{\mathtt{Med}}}
\newcommand{\Lo}{\ensuremath{\mathtt{Low}}}

\begin{document}

\publishingnote{Originally published as Technical report 2014--7. This is an extended version of a paper accepted to \href{http://agi-conf.org/2015/}{AGI-2015}.}

\maketitle

\begin{abstract}
This paper motivates the study of decision theory as necessary for aligning smarter-than-human artificial systems with human interests. We discuss the shortcomings of two standard formulations of decision theory, and demonstrate that they cannot be used to describe an idealized decision procedure suitable for approximation by artificial systems. We then explore the notions of \emph{policy selection} and \emph{logical counterfactuals}, two recent insights into decision theory that point the way toward promising paths for future research.
\end{abstract}

\section{Introduction} \label{sec:intro}

As artificially intelligent machines grow more capable and autonomous, the behavior of their decision procedures becomes increasingly important. This is especially true in systems possessing great general intelligence: superintelligent systems could have a massive impact on the world \citep{Bostrom:2014}, and if a superintelligent system made poor decisions (by human standards) at a critical juncture, the results could be catastrophic \citep{Yudkowsky:2008}. When constructing systems capable of attaining superintelligence, it is important for them to use highly reliable decision procedures.

Verifying that a system works well in test conditions is not sufficient for high confidence. Consider the genetic algorithm of \citet{Bird:2002}, which, if run on a simulated representation of a circuit board, would have evolved an oscillating circuit. Running in reality, the algorithm instead re-purposed the circuit tracks on its motherboard as a makeshift radio to amplify oscillating signals from nearby computers. Smarter-than-human systems acting in reality may encounter situations beyond both the experience and the imagination of the programmers. In order to verify that an intelligent system would make good decisions in the real world, it is important to have a theoretical understanding of why that algorithm, specifically, is expected to make good decisions even in unanticipated scenarios.

What does it mean to ``make good decisions"? To formalize the question, it is necessary to precisely define a process that takes a problem description and identifies the best available decision (with respect to some set of preferences\footnote{For simplicity, assume von Neumann-Morgenstern rational preferences \citep{Von-Neumann:1944}, that is, preferences describable by some utility function. The problems of decision theory arise regardless of how preferences are encoded.}). Such a process could not be \emph{run}, of course; but it would demonstrate a full understanding of the problem of decision-making. If someone cannot formally state what it means to find the best decision in theory, then they are probably not ready to construct heuristics that attempt to find the best decision in practice.

At first glance, formalizing an idealized process which identifies the best decision in theory may seem trivial: iterate over all available actions, calculate the utility that would be attained in expectation if that action were taken, and select the action which maximizes expected utility. But what are the available actions? And what are the counterfactual universes corresponding to what ``would happen" if an action ``were taken"? These questions are more difficult than they may seem.

The difficulty is easiest to illustrate in a deterministic setting. Consider a deterministic decision procedure embedded in a deterministic environment. There is exactly one action that the decision procedure is going to select. What, then, are the actions it ``could have taken"? Identifying this set may not be easy, especially if the line between agent and environment is blurry. (Recall the genetic algorithm repurposing the motherboard as a radio.) However, action identification is not the focus of this paper.

This paper focuses on the problem of evaluating each action \emph{given} the action set. The deterministic algorithm will only take one of the available actions; how is the counterfactual environment constructed, in which a deterministic part of the environment does something it doesn't? Answering this question requires a satisfactory theory of counterfactual reasoning, and that theory does not yet exist.

Many problems are characterized by their idealized solutions, and the problem of decision-making is no exception. To fully describe the problem faced by intelligent agents making decisions, it is necessary to provide an idealized procedure which takes a description of an environment and one of the agents within, and identifies the best action available to that agent. Philosophers have studied candidate procedures for quite some time, under the name of \emph{decision theory}. The investigation of what is now called decision theory stretches back to Pascal and Bernoulli; more recently decision theory has been studied by \citet{Wald:1939}, \citet{Lehmann:1950}, \citet{Jeffrey:1965}, \citet{Lewis:1981}, \citet{Joyce:1999}, \citet{Pearl:2000} and many others.

Various formulations of decision theory correspond to different ways of formalizing counterfactual reasoning. Unfortunately, the standard answers from the literature do not allow for the description of an idealized decision procedure. Two common formulations and their shortcomings are discussed in \Section{counterfactuals}. \Section{susm} argues that these shortcomings imply the need for a better theory of counterfactual reasoning to fully describe the problem that artificially intelligent systems face when selecting actions. Sections~\ref{sec:policies} and~\ref{sec:lcs} discuss two recent insights that give some reason for optimism and point the way toward promising avenues for future research. Nevertheless, \Section{indirect} briefly discusses the pessimistic scenario in which it is not possible to fully formalize the problem of decision-making before the need arises for robust decision-making heuristics. \Section{discussion} concludes by tying this study of decision theory back to the more general problem of aligning smarter-than-human systems with human interests.

\section{Counterfactual Reasoning} \label{sec:counterfactuals}

\subsection{Evidential Decision Theory} \label{sec:edt}

One simple method of counterfactual reasoning, among the first suggested in the decision theory literature \citep[\bibstring{chapter} 5]{Jeffrey:1983}, formalizes the evaluation of what ``would happen" if an agent took action $a$ by taking a Bayesian probability distribution over outcomes and evaluating expected utility conditional on a sentence claiming that the agent takes action $a$. This requires specifying a set $A$ of actions that the agent could take, and the ability to formalize the event ``the agent takes action $a$" for every action $a \in A$.

This sort of counterfactual reasoning, known as ``evidential reasoning," corresponds to the use of evidential decision theory (EDT). Unfortunately, formalizing EDT proves somewhat problematic. Imagine again a deterministic environment. Given a sufficiently accurate description represented as a Bayesian probability distribution over propositions about the environment, the proposition ``the agent takes action~$a$" has probability zero whenever the agent will not, in fact, take action~$a$. As probability distributions conditioned upon events of probability zero are undefined, we cannot use evidential reasoning to identify the best action available to an agent in this case. This can be remedied in part by \emph{inducing} uncertainty about how the agent will act, but this seems unsatisfactory.

A stochastic environment does not much help. Consider, for example, a scenario in which an agent is almost certain to take a suboptimal action, and will only take the best action if hit by cosmic rays. Conditioned on the event ``the agent takes the optimal action," the agent would pay a significant sum to have itself reset (as it has been compromised by cosmic rays). In this scenario, evidential reasoning might mis-identify the suboptimal action as the best available action.

Or, in other words, the problem is not that the distribution conditioned on ``the agent takes action $a$" is sometimes undefined, the problem is that the distribution conditioned on this event may contain spurious correlations (about what could \emph{factually} cause the agent to take action $a$) which are not relevant to the counterfactual.

This is related to another set of concerns with evidential reasoning. EDT is susceptible to what David Lewis refers to as ``an irrational policy of managing the news" \citep{Lewis:1981}, a phenomena further explored by \citet{Arntzenius:2008}. \citet{Eells:1984} has argued that in most cases, EDT can be prevented from managing the news in situations where the agent has sufficient self-knowledge. However, we will demonstrate a scenario in which the defense of Eells does not apply.

\begin{quotation}
\noindent \textbf{The Evidential Blackmail problem.} There is an artificially intelligent agent that plays the stock market. It has amassed substantial wealth. Currently, rumors are circulating about the CEO of one of the companies that the agent has been investing in, and the agent assigns a 0.4\% chance to there being a scandal that forces the CEO to resign; if that is the case then the agent expects to lose 150 million dollars.

    A clever AI researcher, who is renowned for honesty, has access to the source code of the agent. The AI researcher is further known for the ability to predict how AI systems will react in simple thought experiments. For simplicity, assume that the AI researcher is a perfect predictor when given access to an agent's source code. The researcher manages to figure out whether or not there is going to be a scandal (using information unavailable to the agent), and decides to use that knowledge as follows:

    First, the researcher predicts whether or not the agent will pay 100 million dollars after receiving the following message. Then, if either (a) there is \emph{not} a scandal and the researcher predicts that the agent \emph{will} pay, or (b) there \emph{is} a scandal and the researcher predicts that the agent \emph{won't} pay, then the researcher will send the agent a pre-drafted message explaining this whole procedure and that one of either (a) or (b) turned out to be true (without telling the agent which one). The message concludes by asking the agent for 100 million dollars.
\end{quotation}
Evidential reasoning, given full knowledge of this situation, prescribes paying up upon receiving the message. The reasoning runs as follows:
\begin{quote}
    If the agent has received the message, then it must either be the case that it pays and there is not a scandal, or it refuses and there is a scandal. Conditioned on paying, there is not a scandal, which means the agent only loses \$100 million.  However, conditioned on refusing, there is a scandal, and the agent loses \$150 million. The first option loses less money, and so is the best action.
\end{quote}
This reasoning is flawed: the agent's choice of whether or not to pay has no impact upon whether or not the scandal has occurred. If the agent is the type of agent which refuses to pay, then the message is evidence of bad news; but no matter what, paying the researcher results in a needless loss of \$100 million.

However, an agent reasoning according to the prescriptions of EDT would pay the researcher if the message was sent. The researcher, knowing this, would send the message if there was not a scandal, and extract vast sums of money from the agent in this likely case.

Thus, evidential decision theory is not a process that always identifies the best available action. Indeed, this tendency to ``manage the news" by prescribing actions that correlate with (but do not cause) good news, has led to widespread dissatisfaction with evidential decision theory \citep{Lewis:1981,Skyrms:1980,Arntzenius:2008}. A number of people have attempted to patch EDT by giving it a chance to reconsider its decisions partway through the decision process (after learning what it would have done had it not had the opportunity to reconsider) \citep{Jeffrey:1983,Eells:1984,Price:1986,Price:1991}, but these attempts have been largely unsuccessful to date \citep{Ahmed:2005,Egan:2007,Joyce:2007,Ahmed:2010}. Many decision theorists instead prefer methods of counterfactual reasoning that prescribe actions based only upon the causal implications of those actions.

\subsection{Causal Decision Theory} \label{sec:cdt}

A study of counterfactual reasoning which takes into account only what the action causes to happen (directly or indirectly) has led to the development of \emph{causal counterfactual reasoning} and the corresponding causal decision theory (CDT).

Pearl's calculus of interventions on causal graphs \citep{Pearl:2000} can be used to formalize CDT. This requires that the environment be represented by a causal graph in which the agent's action is represented by a single node. This formalization of CDT prescribes evaluating what ``would happen" if the agent took the action $a$ by identifying the agent's action node, cutting the connections between it and its causal ancestors, and setting the output value of that node to be $a$. This is known as a \emph{causal intervention}. The causal implications of setting the action node to $a$ may then be evaluated by propagating this change through the causal graph in order to determine the amount of utility expected from the execution of action $a$. The resulting modified graph is a ``causal counterfactual" constructed from the environment.

\begin{figure}
    \centering
    \begin{tikzpicture}
        \node[circle,draw,minimum size=1cm]
            (S)   at (2.5, 5) {$\mathrm{S}$};
        \node[circle,draw,minimum size=1cm]
            (R)   at (1, 3.5) {$\mathrm{R}$};
        \node[rectangle,draw,minimum size=1cm]
            (A)   at (1, 1.5) {$\mathrm{P}$};
        \node[diamond,draw,minimum size=1cm]
            (U)   at (2.5, 0) {$\mathrm{U}$};
        \draw[->, >=latex] (S) -> (R);
        \draw[->, >=latex] (S) to[out=-45,in=45] (U);
        \draw[->, >=latex] (R) -> (A);
        \draw[->, >=latex] (A) -> (U);
    \end{tikzpicture}
    \caption{The causal graph for the Evidential Blackmail problem. $\mathrm{S}$ denotes whether or not there was a scandal. $\mathrm{R}$ denotes the choice of the AI researcher about whether or not to send the email. $\mathrm{P}$ denotes the choice of the agent about whether or not to pay the researcher. $\mathrm{U}$ denotes the agent's utility, measured in dollars.}
    \label{fig:evidential-blackmail}
\end{figure}
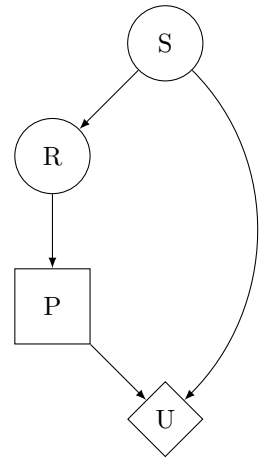

For example, \Figure{evidential-blackmail} details a possible causal graph for the Evidential Blackmail problem. CDT, using this causal graph, would note that the choice whether or not to pay ($\mathrm{P}$) does not impact whether or not there is a scandal ($\mathrm{S}$) and identifies refusal as the best available action.

CDT successfully avoids many of the pitfalls encountered by EDT, and is now the modern academic standard decision theory \citep{Arntzenius:2008,Joyce:2012,Ahmed:2012}. CDT is used under the guise of ``potential outcomes" in statistics \citep{Rubin:1974}, economics \citep[\bibstring{chapter} 1]{Gibbons:1992}, and game theory \citep{Tan:1988}, and implicitly by many modern narrow AI systems under the guise of ``decision networks" \citep[\bibstring{chapter} 16]{Russell:2010}.

Unfortunately, causal counterfactual reasoning is unsatisfactory, for a number of reasons. First and foremost, CDT is underspecified: it is not obvious how to construct a causal graph in which the agent's action is an atomic node. While the environment can be assumed to have causal structure, a sufficiently accurate description of the problem would not represent the agent's action as an ontologically basic entity, but rather as arising from a collection of transistors (or neurons, or sub-atomic particles, etc.). While it seems possible to draw a boundary around some part of the world model which demarcates ``the agent's action," this process may become quite difficult in situations where the line between ``agent" and ``environment" begins to blur, such as scenarios where the agent distributes itself across multiple machines.

It does seem possible to disregard the fact that the agent is composed of parts which follow the laws of physics in order to treat the agent's action as an atomic node. However, while this may be a fine computational expedient for practical decision making, it is not clear that this allows causal decision theory to identify the best action. Consider an environment in which an agent might overheat, and some actions require more computation than others. Can CDT take these side-effects into account, using a model that has thrown away information about individual transistors? It seems plausible that graphical world-models could capture decision problems of this form, but it is not yet obvious how to do so.

Even given a satisfactory graphical model of the environment, causal counterfactual reasoning itself is unsatisfactory. CDT prescribes low-scoring actions on a broad class of decision problems where high scores are possible, known as \emph{Newcomblike problems} \citep{Nozick:1969}. The problem can be exemplified in the following setting:

Consider a one-shot Prisoner's Dilemma played by two identical deterministic agents. Each agent knows that the other is identical. Agents must choose whether to cooperate (C) or defect (D) without prior coordination or communication, with payoffs as in \Table{pd}.\footnote{This scenario (and other Newcomblike scenarios) are multi-agent scenarios. Why use decision theory rather than game theory to evaluate them? The goal is to define a procedure which reliably identifies the best available action; the label of ``decision theory" is secondary. The desired procedure must identify the best action in all settings, even when there is no clear demarcation between ``agent" and ``environment." Game theory informs, but does not define, this area of research.}

\bgroup
\def\arraystretch{1.5}
\begin{table}
    \centering
    \begin{tabular}{l|c|c|}
        \multicolumn{1}{r}{} & \multicolumn{1}{c}{C} & \multicolumn{1}{c}{D} \\
        \cline{2-3}
        C & (2, 2) & (0, 3) \\
        \cline{2-3}
        D & (3, 0) & (1, 1) \\
        \cline{2-3}
    \end{tabular}
    \caption{\label{tab:pd}The Prisoner's Dilemma.}
\end{table}
\egroup

\begin{figure}
    \centering
    \begin{tikzpicture}
        \node[rectangle,draw,minimum size=1cm] (nAgent)   at (1, 3) {$\mathrm{A}$};
        \node[circle,draw,minimum size=1cm] (nOpponent) at (5, 3) {$\mathrm{O}$};
        \node[diamond,draw] (nUtil)   at (3, 1) {$\mathcal{U}$};
        \draw[->, >=latex] (nAgent) -> (nUtil);
        \draw[->, >=latex] (nOpponent) -> (nUtil);
    \end{tikzpicture}
    \caption{The causal graph for a one-shot Prisoner's Dilemma. $\mathrm{A}$
        is the node representing the agent's action, $\mathrm{O}$ is the node
        representing the opponent's action, and $\mathcal{U}$ is the node
        representing the agent's utility.}
    \label{fig:cdt-pd}
\end{figure}
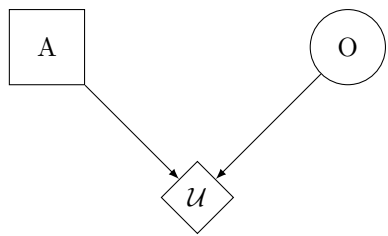

The actions of the two agents will be identical by assumption, but neither agent's action causally impacts the others': in a causal model of the situation, the action nodes are causally separated, as in \Figure{cdt-pd}. When determining the best action available to the left agent, a causal intervention changes the left action node without affecting the right action node, assuming there is some (fixed) probability $p$ that the right agent will cooperate \emph{independent} of the left agent. No matter what the value of $p$ is, CDT reasons that the left agent gets utility $2p$ if it cooperates and $2p + 1$ if it defects, and therefore prescribes defection \citep{Lewis:1979,Joyce:1999}.

Indeed, many decision theorists hold that it is in fact rational for an agent to defect against a perfect copy of itself in a one-shot Prisoner's Dilemma, as after all, no matter what the opponent does, the agent does better by defecting \citep{Gibbard:1978,Lewis:1981,Joyce:2002}. Others object to this view, claiming that since the agents are identical, both actions must match, and mutual cooperation is preferred to mutual defection, so cooperation is the best available action \citep{Bar-Hillel:1972,Altair:2013}. The prescriptions of CDT are not entirely unsatisfactory: if we have an opportunity to change the code of the agent after it is copied but before it acts, then CDT correctly identifies that the agent should be reprogrammed to defect. But if instead we are writing the decision algorithm of an agent that will be played against a perfect copy of itself, then it should \emph{not} follow the prescriptions of CDT. In the moment, CDT identifies defection as the best available option. But this is incorrect: if you ever find yourself in a one-shot Prisoner's Dilemma against an opponent guaranteed to act identically, then you would do better to cooperate.\footnote{It is common to object that reality is stochastic and people seldom encounter identical copies of themselves, and therefore these situations do not matter. We note both that (1) the opponent need not be a perfect copy; it seems prudent to cooperate in a Prisoner's Dilemma against an opponent guaranteed to take the same action as you 90\% of the time and the opposite action 10\% of the time; and (2) the goal is a formally specified means of identifying the best available decision, and for this it is useful to explore edge cases.}

CDT assumes it can hold the action of one opponent constant while freely changing the action of the other, because the actions are causally separated. However, the actions of the two agents are \emph{logically} connected; it is impossible for one agent to cooperate while the other defects. Causal counterfactual reasoning neglects non-causal logical constraints.

Roughly speaking, these non-causal logical constraints arise whenever parts of the environment are logically correlated with (but not causally connected to) the agent's action. This can happen readily in any environment where other actors possess information about the agent's decision-making procedure and are basing their action on the agent's expected action.

It is a common misconception that Newcomblike scenarios only arise when some other actor is a perfect predictor (perhaps by being an identical copy). This is not the case: while Newcomblike scenarios are most vividly exemplified by situations involving perfect predictors, they can also arise when other actors have only partial ability to predict the agent \citep{Lewis:1981a,Ahmed:2014a,Ahmed:2014b}. For example, consider a situation in which an artificial agent is interacting with its programmers, who have intimate knowledge of the agent's inner workings. The agent could well find itself embroiled in a Prisoner's Dilemma with its programmers. Let us assume that the agent knows the programmers will be able to predict whether or not it will cooperate with 90\% accuracy. In this case, even though the programmers are imperfect predictors, the agent is in a Newcomblike scenario.

As with EDT, the tendency of CDT to misidentify suboptimal decisions as the best decision is a flaw that could be used to exploit agents reasoning according to the prescriptions of CDT, as in the following scenario:
\begin{quotation}
\noindent \textbf{The Counterfactual Blackmail problem.} There is an artificially intelligent agent that plays the stock market. It's quite competent, and has amassed substantial wealth. A clever AI researcher, renowned for honesty and for the ability to predict the behavior of AI systems in simple thought experiments, acquires the source code of an artificial agent. For simplicity, assume that the researcher is a perfect predictor when in possession an agent's source code.

    The researcher has developed a computer virus which will affect market operations and cause a massive crash. If the virus is used, both the researcher and the agent will lose \$150 million at least. The virus is designed so that after being deployed it will remain deactivated for a day, such that the only way to prevent activation is by the agent wiring \$100 million to the researcher within 24 hours. If the researcher decides to deploy the virus, they would then send a message to the agent demonstrating that the virus has been deployed and demanding \$100 million.

    The researcher is very risk averse, and will only deploy the virus upon becoming quite sure that the agent will in fact pay up to deactivate it.
\end{quotation}

CDT, evaluating this environment in the situation where the virus has been deployed, prescribes paying up as the best action: for once the virus has been deployed, the agent loses \$100 million if it pays and \$150 million or more if it doesn't. Thus, an agent reasoning according to the prescriptions of CDT at the time of the blackmailing would pay up, and the AI researcher (seeing that the agent reasons according to CDT) would deploy the virus, extracting large sums of money from the agent. This behavior seems unsatisfactory: if the agent were to refuse to pay, then the blackmailer would never deploy the virus in the first place; this results in a better outcome.

As in the case of EDT, a number of people have attempted to patch CDT by giving the agent an opportunity to reconsider partway through the decision process (after learning what it would have done if it hadn't been given the opportunity to reconsider) \citep{Joyce:2007,Arntzenius:2008,Gustafsson:2011}, but the results have largely been unsatisfactory \citep{Egan:2007,Wedgwood:2013}.

The goal is to formalize what is meant when asking that agents take ``the best available action." Causal decision theory often identifies the best action available to an agent, but it sometimes fails in dangerous ways.

\section{Optimization Targets} \label{sec:susm}

What is meant by a ``good decision"? \Section{counterfactuals} demonstrated that neither evidential nor causal formulations of decision theory characterize a satisfactory answer to this question. A better understanding of the question is needed before attempting to evaluate whether or not a given practical heuristic constitutes an answer.

Here an objection arises: why must the programmers attain a better understanding of decision theory? Why can't this problem be delegated to the system itself? Wouldn't the agent have incentives to improve its own decision making procedure? Surely, an intelligent system would be able to identify and address systematic failures in its own decision processes. After all, the question of what procedure to use for making decisions is itself a decision problem \citep{Skyrms:1982}.

Indeed, an intelligent system would generally have incentives to improve upon its decision-making heuristics \citep{Omohundro:2008}, and self-modifying agents acting according to the prescriptions of CDT or EDT would be able to eliminate much of the suboptimal behavior described in \Section{counterfactuals}. However, such systems would not by default converge on satisfactory decision procedures, for reasons that we will now explain.

To illustrate the problem, it is useful to imagine hypothetical ``CDT agents" choosing according to the prescriptions of CDT. Such an agent is impractical---CDT assumes total world-knowledge and fully evaluates the expected utility of all possible actions---but it serves to demonstrate that practical decision heuristics approximating causal reasoning would be approximating unsatisfactory behavior.

Consider an environment containing a self-modifying CDT agent that is about to be copied, at which point it will face that copy in a one-shot Prisoner's Dilemma. The agent has the opportunity to self-modify and change its decision procedure. What is the prescription of CDT? If the agent continues to act according to CDT after it is copied, it will defect and achieve payoff 1. But if it self-modifies to unilaterally cooperate with its copy, then it will cooperate and achieve payoff 2. Thus, before the agent is copied, CDT prescribes self-modifying to stop using CDT.

This result generalizes: if an agent is about to face a situation where CDT performs poorly, then CDT correctly identifies that the agent should modify itself to stop using CDT to make decisions. (We say that CDT is ``unstable under reflection," or ``not self-recommending," to use the terminology of  \citet{Skyrms:1982} and \citet{Meacham:2010}.) This gives rise to the question of what decision theory CDT does prescribe using: given an environment in which an agent may self-modify to adopt any approach to decision theory, which does CDT identify as the best available option? If this selection were satisfactory, then it would not matter that CDT itself is unsatisfactory, as self-modifying agents reasoning according to CDT would converge on good decision-making procedures.

Unfortunately, however, the answer is not satisfactory. In fact, CDT prescribes that an agent \emph{resist} certain attempts to improve its decision procedures.

Consider again the scenario where a self-modifying agent will face a copy of itself in the Prisoner's Dilemma. Assume the scenario works as follows: (1) at time $t=1$ the agent has an opportunity to self-modify; (2) at time $t=2$ the agent is copied; (3) at time $t=3$ each agent has the opportunity to self-modify; (4) at time $t=4$ each agent is pitted against the other in a Prisoner's Dilemma. Assuming the agent is a CDT agent at time $t=1$, CDT prescribes self-modifying to cooperate, as this leads to higher expected utility. But if the agent fails to self-modify at $t=1$, then what is the prescription of CDT at $t=3$? In this case CDT prescribes no self-modification, because it considers that the copies are causally independent, neglecting the logical fact that if one self-modifies then so will the other.

At $t=1$, CDT prescribes cooperation, but at $t=3$ it prescribes defection. Yet, in all times, CDT is attempting to identify the best available action according to the same set of preferences.

\emph{Expected utility maximization alone is not sufficient to determine the best available action!} It is the combination of expected utility maximization with a method for formalizing counterfactuals which defines the best available action.

We refer to this combination of preferences with counterfactuals as an	 \emph{optimization target}. At $t=1$, the optimization target of CDT is the maximization of the agent's utility according to a causal counterfactual constructed at $t=1$. At $t=3$, the optimization target has shifted: the preferences are the same, but now the maximization is done according to a causal counterfactual constructed at $t=3$.

The optimization target of CDT drifts over time: $\text{CDT}_{t=1}$ reasons that an agent acting according to $\text{CDT}_{t=3}$ would choose the best action according to the optimization target of $\text{CDT}_{t=3}$, and this is different from the optimization target of $\text{CDT}_{t=1}$. CDT prescribes preventing this drift. \mkbibparens{Another way of saying this is that CDT prescribes paying for the opportunity to precommit to certain actions, as explained by \citet{Burgess:2004}.}

This answers the question of what decision theory CDT would prescribe that a self-modifying agent select: at time $t$, CDT prescribes that an agent self-modify to forevermore optimize according to what would have been the best action according to a causal counterfactual constructed at time $t$. An agent reasoning according to a decision theory of this form would outperform an agent reasoning according to unmodified CDT: for example, it would attain high scores in all Newcomblike problems which began after time $t$. However, it would continue to preform poorly in all Newcomblike scenarios that began before time $t$.

This may seem acceptable: what does it matter if the agent scores poorly in Newcomblike scenarios that started in its past, so long as it can succeed in Newcomblike scenarios that begin in its future? Unfortunately, it may matter quite a bit: any actor with access to the agent's original source code would have the ability to put the agent into a Newcomblike scenario that \emph{began in the agent's causal past.}

To demonstrate, consider the following scenario, which is identical to the counterfactual blackmail problem except that the AI researcher now has access to a copy of the agent's \emph{original} source code, and the agent has the opportunity to self-modify after the researcher steals the source code but before the researcher decides whether or not to deploy their virus:

\begin{quotation}
    \noindent \textbf{The Retro Blackmail problem.} There is a wealthy intelligent system and an honest AI researcher with access to the agent's original source code. The researcher may deploy a virus that will cause \$150 million each in damages to both the AI system and the researcher, and which may only be deactivated if the agent pays the researcher \$100 million. The researcher is risk-averse and only deploys the virus upon becoming confident that the agent will pay up.

    The agent knows the situation and has an opportunity to self-modify after the researcher acquires its original source code but before the researcher decides whether or not to deploy the virus. (The researcher knows this, and has to factor this into their prediction.)
\end{quotation}

Clearly, it is preferable for the agent use its opportunity to self-modify in order to ensure that it would not give in to the demands of the rogue researcher. The researcher, then, would be able to deduce from the agent's original source that it is the type of agent which would self-modify in order to precommit to refusing the rogue's demands, and would not deploy the virus.

However, according to a causal counterfactual constructed any time at or after the agent's inception, the behavior of the ``copy" that the researcher reasons about (by inspecting the original source code) is causally disconnected from the behavior of the agent: CDT and \emph{any decision procedure to which CDT would self-modify} see losing money to the blackmailer as the best available action.

To illustrate what went wrong, let us anthropomorphize the reasoning of an agent which started out as a CDT agent. This agent would never self-modify to address the vulnerability, even if given the opportunity to do so before the researcher accessed its original source. Its refusal to precommit might be justified as follows:

\begin{quote}
    I am not being retro-blackmailed now, but I may be in the future. Consider a self-modification designed to prevent retro-blackmail by precommitting now to refuse all future retro-blackmailer demands. Retro-blackmailers decide whether or not to blackmail according to the predicted behavior of an ``instance" of me spawned from my \emph{original} source code. My decision to precommit now does \emph{not} control the choice of that instance, as its choices are independent of my own. If that instance were to decide to precommit to refuse all retro-blackmailer demands, then it doesn't matter what I choose now. But if that instance decided not to precommit, then I had better not make the precommitment, because I'll still be blackmailed and it's better to give in once blackmailed.  Therefore, I won't apply the self-modification.
\end{quote}
By contrast, we would like to understand and formalize the decision theory that corresponds to arguments like the following:
\begin{quote}
    I am not being retro-blackmailed now, but I may be in the future. Consider a self-modification designed to prevent retro-blackmail by precommitting now to refuse all future retro-blackmailer demands. Retro-blackmailers decide whether or not to deploy their viruses according to the actions of an ``instance" of me spawned from my \emph{original} source code, which, being an instance of me, will also consider self-modifying to avoid future retro-blackmail and which will come to the same conclusion as I do right now. If I apply the self-modification, then so will it, and people looking at my original source would predict this, so I would never be retro-blackmailed. But if I don't apply the self-modification, then nor will that instance, and people looking at my original source would predict that I didn't, and I'd be retro-blackmailed. The first option is clearly preferable, so I apply the self-modification.
\end{quote}
However, CDT neither prescribes such a modification nor the adoption of any decision procedure which would accept such a modification. Indeed, a CDT agent that knows it is about to be retro-blackmailed might pay a significant sum to \emph{avoid} the patch.\footnote{The specific amount it would be willing to pay depends upon the probability with which it believes its simulated copy would avoid the patch. This could consistently be 1, depending how the causal graph is constructed; in this case the agent would be willing to pay \$49 million to prevent the patch.}

We have shown that self-modifying systems approximating CDT would not converge on a satisfactory decision procedure. While intelligent agents have incentives to improve their ability to hit their current optimization target, they would not by default change their original optimization targets. When constructing practical decision-making systems, ensuring that they are deciding according to the right preferences is not enough; one must also ensure that they make decisions according to a satisfactory optimization target.

The optimization target of CDT is unsatisfactory. Evidential decision theory suffers from similar instability under self-modification, for similar reasons. Satisfactory optimization targets are not yet well understood.

Lest the reader grow pessimistic from all this discussion of why the problem is difficult, the following two sections introduce two ideas leading to a new formalization of decision theory that has a better optimization target and avoids many of the shortcomings of both CDT and EDT.

\section{Policy Selection} \label{sec:policies}

Consider the reasoning that \emph{humans} might use to determine that it is better to refuse to pay up on the (normal, not retro) counterfactual blackmail problem:

\begin{quote}
    Consider an agent that would pay up in response to a counterfactual blackmail. The blackmailer would predict this and blackmail the agent. Now, instead, consider an agent that would refuse to pay up in response to a counterfactual blackmail. The blackmailer would predict this too, and so would not blackmail the agent. Therefore, if we are constructing an agent that might encounter counterfactual blackmail, then it is a better overall policy to construct an agent that would refuse to pay up when blackmailed in this way.
\end{quote}
Notice that we are choosing between \emph{global policies} of ``always pay" or ``never pay," without regard for the agent's specific observations. Compare this to the anthropomorphized reasoning of a CDT agent deciding whether or not to pay up to a counterfactual blackmail, upon seeing that the virus has been deployed:
\begin{quote}
    Alas, the virus has been deployed. I would have preferred that the virus not be deployed, but since it has been, I must now decide whether or not to pay up. Paying up is bad, but refusing is worse, so I'll pay up.
\end{quote}

If you ask a person what the agent should do upon observing the blackmail, they might say ``ah, the answer to this question determines whether or not the agent will be blackmailed; therefore the agent should not pay." By contrast, if you hand CDT a description of the environment in which the virus has been deployed, it will prescribe paying: because, alas, the virus has already been deployed. (Notice, however, that CDT also chooses as we do when deciding what type of agent it would prefer to place into a counterfactual blackmail scenario. That is, while CDT prescribes paying, it prescribes constructing agents which refuse.)

This demonstrates two different methods by which agents can select actions. We, on the one hand, prefer to construct agents which apply the optimal \emph{observation-to-action mapping}.\footnote{Earlier, we mentioned the problem of identifying the set of actions available to the agent. That problem is beyond the scope of this paper; we have been assuming that the set of available actions is given. Now we must also assume that there is some way of identifying the agent's \emph{observations}, such that the set of observations is also given; an observation-to-action mapping is then any function which maps observations to actions.} A map from observations is known as a \emph{policy}, and a decision theory which iterates over policies (rather than actions) is said to use ``policy selection."

The alternative is to compute the best available action given observation. Oftentimes, these two methods are equivalent; conditioning on observation is usually the best policy. However, whenever the prescriptions of these different methods differ, agents using policy selection outperform agents that condition on observation: an agent using policy selection can always act as if it's conditioning on all of its observations, but an agent conditioning on its observations cannot always act as if it is using policy selection.

CDT implicitly corresponds to the latter (as does EDT, for that matter). There is a sense in which CDT ``prefers" policy selection instead \citereset\citep{Burgess:2004}; it prescribes that an agent pay for the opportunity to precommit to refusing blackmail if the agent's source code is about to be stolen. However, once the observation of blackmail is made, it prescribes paying: it does not have agents act as it would have precommitted to act.

Given that it is better for agents to act as they would have precommitted to act, it is possible to improve upon CDT by simply specifying a variant of CDT that evaluates actions according to the best available policy. This can be done by specifying a version of CDT that, instead of iterating over actions available to the agent and selecting the best one, iterates over \emph{observation-to-action mappings}, selects the best, and then applies the agent's observations to identify the best action.

For example, in the counterfactual blackmail problem, an agent reasoning according to policy selection might reason along the following lines:
\begin{quote}
    The optimal policy is to refuse to pay up upon observing that the virus has been deployed. I now observe that the virus has been deployed. Therefore, I refuse to pay.
\end{quote}
It may seem strange for the agent to refuse to condition upon its own observation when selecting its action, but this is necessary in order for the agent to succeed. In a sense, the agent is acting as if it can't tell whether it is actually observing the virus or whether it is merely the blackmailer's prediction, because, after all, the blackmailer is reasoning about what the agent \emph{would} do if the agent \emph{did} observe the deployment of the virus.

A decision theory that identifies the best \emph{policy} in a given scenario (and prescribes acting accordingly) better captures the notion of ``the best available action" than a decision theory which considers actions alone. Variants of decision theory using policy selection are ``updateless" (as agents following the prescriptions of policy selection pick a policy before they update on their observations), and this is the first of two ideas behind the \emph{updateless decision theory} (UDT) of \citet{Dai:2009}. This idea follows in the wake of \citet{Gauthier:1994}, who advocated making decisions using global policy selection, and \citet{Arntzenius:2004}, who applied this idea to an infinite decision problem similar to the ``Procrastination Paradox" of \citet{Yudkowsky:2013c}. Another decision procedure similar to that of Dai was proposed by \citet{Meacham:2010}.

Policy selection is essentially what CDT would prescribe in order to prevent optimization target drift: an updateless variant of CDT correctly identifies the best action in the counterfactual blackmail problem. However, this variant of CDT still has a bad optimization target: it still prescribes paying up in the retro blackmail problem. To address that shortcoming, the second insight of UDT is necessary.

\section{Logical Counterfactuals} \label{sec:lcs}

Consider the sort of reasoning that a human might use, faced with a Prisoner's Dilemma in which the opponent's action is guaranteed to match our own:
\begin{quote}
    The opponent will certainly take the same action as I take. Thus, there is no way for me to exploit the opponent, and no way for the opponent to exploit me. Either we both cooperate and I get \$2, or we both defect and I get utility \$1. I prefer the former, so I cooperate.
\end{quote}
Contrast this with the hypothetical reasoning of a reasoner who, instead, reasons according to causal counterfactuals:
\begin{quote}
    There is some probability $p$ that the opponent defects. Consider cooperating. In this case, I get \$2 if the agent cooperates and \$0 otherwise, for a total of \$2$p$. Now consider defecting. In this case I get \$3 if the opponent cooperates and \$1 otherwise, for a total of \$2$p$ $+$ 1. Defection is better no matter what value $p$ takes on, so I defect.
\end{quote}
We reason as if our decision controls both agents; the causal reasoner assumes that their action is independent from the action of the opponent.

More generally, identifying the best action requires respecting the fact that identical algorithms produce identical outputs. However, CDT evaluates actions according to a \emph{physical} counterfactual where the action is changed but everything causally separated from the action is held constant.

It is not the \emph{physical output of the agent's hardware} which must be modified to construct a counterfactual, it is the \emph{logical output of the agent's decision algorithm}. This is the second insight \mkbibparens{discovered independently by \citet{Yudkowsky:2010} and by \citet{Spohn:2012}} behind Wei Dai's UDT.

To give an intuition for what this entails, consider a motivating example. In the symmetric Prisoner's Dilemma, CDT identifies the best action according to a causal model of the world, as in \Figure{cdt-pd}. CDT reasons that the agent's action is causally disconnected from the opponent's action. CDT considers each action according to a counterfactual in which the agent's action is changed while the opponent's action is held constant. This leads to the consideration of impossible scenarios where one agent cooperates while the other defects, and this in turn leads to the conclusion that defection is a dominant strategy, so CDT misidentifies ``defect" as the best available action.

By contrast, UDT counterfactually considers what would happen if an agent selected the action $a$ by constructing a counterfactual in which \emph{that algorithm} outputs $a$. Thus, when considering what would happen if the agent defects, it evaluates a counterfactual world in which both the agent and the opponent defect.

\begin{figure}
    \centering
    \begin{tikzpicture}
        \node[rectangle,draw,minimum size=1cm] (nUDT) at (3, 5) {\Algo};
        \node[circle,draw,minimum size=1cm] (nAgent)   at (1, 3) {$\mathrm{A}$};
        \node[circle,draw,minimum size=1cm] (nOpponent) at (5, 3) {$\mathrm{O}$};
        \node[diamond,draw] (nUtil)   at (3, 1) {$\mathcal{U}$};
        \draw[->, >=latex] (nUDT) -> (nAgent);
        \draw[->, >=latex] (nUDT) -> (nOpponent);
        \draw[->, >=latex] (nAgent) -> (nUtil);
        \draw[->, >=latex] (nOpponent) -> (nUtil);
    \end{tikzpicture}
    \caption{The logical graph for a symmetric Prisoner's Dilemma where both the agent's action $\mathrm{A}$ and the opponent's action $\mathrm{O}$ are determined by the algorithm \Algo.}
    \label{fig:udt-pd}
\end{figure}
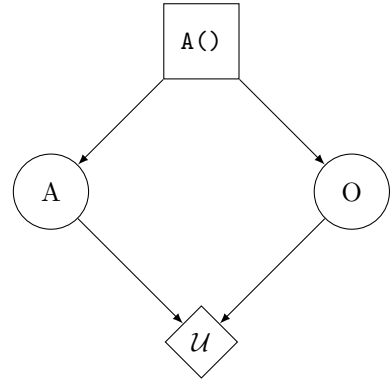

More generally, UDT chooses the best action according to a world-model which represents not only causal relationships in the world, but also the logical effects of algorithms upon the world. In the symmetric Prisoner's Dilemma, UDT may reason according to a world-model that looks something like \Figure{udt-pd}, in which the output of each agent is determined by the same algorithm.

By way of illustration, let us anthropomorphize the reasoning of an agent acting according to the prescriptions of UDT in the real world:
\begin{quote}
The physical actions of both myself and my opponent are determined by the same algorithm. Therefore, whatever action this very decision algorithm selects will be executed by both of us. If this decision algorithm selects ``cooperate" then we'll both cooperate and I'll get a payoff of $2$. If instead this decision algorithm selects ``defect" then we'll both defect and I'll get a payoff of $1$. Therefore, I (this decision algorithm) select ``cooperate."
\end{quote}
Using reasoning of this form, a selfish agent acting according to the prescriptions of UDT cooperates with an identical agent on a symmetric one-shot Prisoner's Dilemma, and achieves the higher payoff.\footnote{Note that the agent does \emph{not} care about the utility of its opponent. Each agent is maximizing its own individual utility. Both players understand that the payoff must be symmetric, and cooperate out of a selfish desire to achieve the higher symmetric payoff.}

In order to identify the best available action, it is important to respect the logical connections in the environment, not only the causal connections. When considering a counterfactual in which an agent selects a given action, it is important to construct a counterfactual in which \emph{all instances of the agent's decision process} select that action; otherwise the logical connections in the environment are destroyed.

While this idea sounds promising, it has proven difficult to formalize. Two partial attempts at formalizing UDT are detailed below, but no fully satisfactory formalization of ``logical counterfactuals" yet exists.

\subsection{First Attempt: Graphical UDT} \label{sec:gudt}

As alluded to by \Figure{udt-pd} above, UDT can be formalized using a graphical approach similar to Pearl's formalization of CDT \citeyearpar{Pearl:2000}. To do so, the graphical representation of the environment must encode not only causal relations, but also ``logical relations." Unfortunately, it is not yet entirely clear how to encode ``logical relations" in a graph, nor how updates should be propagated through the graph after intervening on one of the nodes.

Given a probabilistic graphical model of the world representing both logical and causal connections, and given that one of the nodes in the graph corresponds to the agent's decision algorithm, and given some method of propagating updates through the graph, UDT can be specified in a manner very similar to CDT. To identify the best policy available to an agent, iterate over all available policies $\pi \in \Pi$, change the value of the agent's algorithm node in the graph to $\pi$, propagate the update, record the resulting expected utility, and return the policy $\pi$ leading to the highest expected utility.\footnote{The best action, then, is $\pi(o)$ where $o$ is the agent's observation, as per \Section{policies}. In this paper, we leave aside issues of formalizing what counts as an ``observation," just as we leave aside the issue of identifying the set of ``possible actions."} Of course, the difficult part in this process is the construction of a graph representing the ``appropriate" logical and causal connections, which is somehow capable of propagating changes in a satisfactory manner. It is not at all clear how to construct such a graph given an arbitrarily accurate world model composed of molecules rather than logical algorithms, for reasons similar to those discussed in \Section{cdt}.

In other words, UDT (like CDT) is underspecified, pending a formal description of how to construct such a graph. However, constructing a graph suitable for UDT is significantly more difficult than constructing a graph suitable for CDT. While both require decreasing the resolution of the world model until the agent's action (in CDT's case) or algorithm (in UDT's case) is represented by a single node rather than a collection of parts, the graph for UDT further requires some ability to identify and separate ``algorithms" from the physical processes that implement them. How is UDT supposed to recognize that the agent and its opponent implement the same algorithm? Will this recognition still work if the opponent's algorithm is written in a foreign programming language, or otherwise obfuscated in some way? Successfully identifying all copies of an algorithm in the world is no easy feat.\footnote{Something vaguely similar, however, has been shown to be possible in certain restricted scenarios \citep{LaVictoire:2014}.}

\begin{figure}
    \centering
    \begin{tikzpicture}
        \node[rectangle,draw,minimum size=1cm] (nUDT) at (3, 5) {\Algo};
        \node[circle,draw] (nRandom) at (6, 4) {$X$};
        \node[circle,draw,minimum size=1cm] (nAgent)   at (1, 3) {$\mathrm{A}$};
        \node[circle,draw,minimum size=1cm] (nOpponent) at (5, 3) {$\mathrm{O}$};
        \node[diamond,draw] (nUtil)   at (3, 1) {$\mathcal{U}$};
        \draw[->, >=latex] (nUDT) -> (nAgent);
        \draw[->, >=latex] (nUDT) -> (nOpponent);
        \draw[->, >=latex] (nRandom) -> (nOpponent);
        \draw[->, >=latex] (nAgent) -> (nUtil);
        \draw[->, >=latex] (nOpponent) -> (nUtil);
    \end{tikzpicture}
    \caption{The desired logical graph for the one-shot Prisoner's Dilemma where agent $\mathrm{A}$ acts according to \Algo, and the opponent either mirrors \Algo\ or does the opposite, according to the random variable $X$.}
    \label{fig:udt-rpd}
\end{figure}
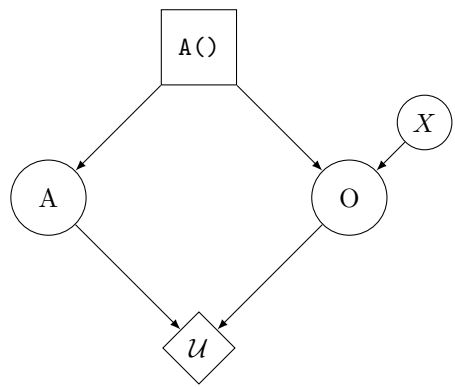

Even given some reliable means of identifying copies of an agent's decision algorithm in the environment, this may not be enough to specify a satisfactory graph-based version of UDT. To illustrate, consider UDT identifying the best action available to an agent playing a Prisoner's Dilemma against an opponent that does exactly the same thing as the agent 80\% of the time, and takes the opposite action otherwise. It seems UDT should reason according to a graph as in \Figure{udt-rpd}, in which the opponent's action is modeled as dependent both upon the agent's algorithm and upon some source $X$ of randomness. However, generating logical graphs as in \Figure{udt-rpd} is a more difficult task than simply detecting all perfect copies of the an algorithm in an environment.

More generally, a satisfactory formalization of graphical UDT must account for the logical connection between the agent's algorithm and all other algorithms. Clearly, under the assumption that the agent's algorithm selects the policy $\pi$, the algorithm ``do what the agent does 80\% of the time" is affected. But what about other algorithms which correlate with the agent's algorithm, despite not referencing it directly? What about the decision algorithms of other agents which base their decisions on an imperfect model of how the agent will behave? A satisfactory formalization of the ``logical implications" of the agent selecting $\pi$ must give an account of how this logical counterfactual affects all other algorithms, and this seems difficult to do in general.

Given all of these difficulties, there is not yet a satisfactory formalization of graph-based UDT. However, an alternative formalization, known as ``proof-based UDT," addresses many of these concerns and provides a more complete (although still ultimately unsatisfactory) formalization.

\subsection{Second Attempt: Proof-Based UDT} \label{sec:pudt}

UDT prescribes reasoning about the effects of taking policy $\pi$ by evaluating the logical implications of the agent's algorithm selecting the policy $\pi$. As discussed above, a satisfactory description of the logical implications of $\Algo=\pi$ requires some method of reasoning about how this assumption impacts all other algorithms.

Given some method of reasoning about the effects of $\Algo=\pi$ on any other algorithm, though, a graphical formalization of UDT is unnecessary: \emph{the environment itself is an algorithm}, which describes how to compute the agent's expected utility!

That is, a formal understanding of ``logical implication" could be leveraged to analyze the effects of $\Algo=\pi$ upon the environment. Thus, to evaluate the policy $\pi$, UDT need only compute the expected utility available in the environment as modified by the assumption $\Algo=\pi$.

This realization leads to the idea of ``proof-based UDT," which evaluates policies by searching for formal proofs, using some mathematical theory such as Peano Arithmetic ($\PA$), of how much utility is attained in the world-model if $\Algo$ selects the policy $\pi$. As a bonus, this generic search for formal proofs obviates the need to identify the agent in the environment: given an environment which embeds the agent and a description of the agent's algorithm, then no matter how the agent is embedded in the environment, there will be some formal proof which identifies it and describes the implications of that algorithm outputting $\pi$. While that proof must do the hard work of identifying all copies and variants of $\Algo$ and describing how it interacts with the environment, the high-level UDT algorithm simply searches all proofs, with no need for formalizing some way of locating the agent. This allows for an incredibly simple specification of updateless decision theory, given below.

First, a note on syntax: Square quotes ($\Quote{\,\cdot\,}$) denote sentences encoded as objects that a proof searcher can search for. This may be done via e.g.\ a G\"odel encoding. Overlines within quotes denote ``dequotes," allowing the reference of meta-level variables. That is, if at some point in the algorithm $\pi := 3$ and $o := 10$, then the string $\Quote{\Algo=\overline{\pi}\to \Env=\overline{o}}$ is an abbreviation of $\Quote{\Algo=3 \to \Env=10}$. The arrow $\Quote{\to}$ denotes logical implication.

The algorithm is defined in terms of a finite set $\Pi$ of policies available to the agent and a finite sorted list $O$ of outcomes that could be achieved (ordered from best to worst). The proof-based UDT algorithm takes a description $\Quote{\Env}$ of the environment and $\Quote{\Algo}$ of the agent's algorithm. $\Env$ computes an outcome, $\Algo$ computes a policy. It is assumed (but not necessary) that changing the output of $\Algo$ would change the output of $\Env$.

\begin{algorithm}
\SetKwFunction{UDT}{UDT}
\SetKwProg{Fn}{Function}{:}{end}
\Fn{\UDT{\Quote{\Env}, \Quote{\Algo}}}{
  \For{outcome $o \in O$}{
    \For{policy $\pi \in \Pi$}{
      \If{$\PA \text{\ proves\ } \Quote{\Algo=\overline{\pi} \to \Env=\overline{o}}$}{
        \KwRet{$\pi$}
      }
    }
  }
  \KwRet{the lexicographically first policy in $\Pi$}
}
\caption{Proof-based UDT\label{alg:pbudt}}
\end{algorithm}

To demonstrate how the algorithm works, consider UDT evaluating the actions of an agent faced with a retro-blackmail scenario. For simplicity, assume the agent is robbed of its chance to self-modify after the AI researcher steals its original source; this opportunity is not necessary in order for UDT to identify the best solution. The list of outcomes is $O := [\,0, -100, -150\,]$ according to the cases where (a) the agent is not blackmailed, (b) the agent is blackmailed and pays up, and (c) the agent is blackmailed and refuses to pay up, respectively.  The set of policies is $\Pi := \{\,\Pay, \Refuse\,\}$ according to whether the agent pays or refuses. To identify the best action, UDT iterates over outcomes in order of preference, starting with $0$. For each outcome, it iterates over policies; say it first considers $\Pay$. In the case that $\Algo=\Pay$, the agent is blackmailed, and so it does not achieve the outcome $0$, so there is no proof of $\Quote{\Algo=\Pay \to \Env=0}$. Next, UDT considers $\Refuse$. In the case that $\Algo=\Refuse$, the instance of $\Algo$ instantiated by the blackmailer would also refuse to pay, and so the agent would never be blackmailed. Therefore, there is a proof of $\Quote{\Algo=\Refuse \to \Env=0}$, and so UDT selects $\Refuse$.

While this proof-based formalism of UDT is extremely powerful, it is not without its drawbacks. It requires a halting oracle in order to check whether proofs of the statement $\Quote{\Algo=\overline{\pi} \to \Env=\overline{o}}$ exist; but this is forgivable, as it is meant to be a definition of what it means to ``choose the best policy," not a practical algorithm. However, this formalization of UDT can only identify the best policy if there exists a proof that executing that policy leads to a good outcome. This is problematic in stochastic environments, and in any setting where $\PA$ is not a strong enough theory to find the appropriate proofs, which may well occur if agents in the environment are themselves searching for proofs about what UDT will prescribe (in order to guess the behavior of agents which act according to UDT).

Even if UDT can find proofs for every policy, there are environments in which UDT still misidentifies the best policy. For example, consider a simple two-player game, described by \citet{Slepnev:2011}, played by a human and an agent which is capable of fully simulating the human and which acts according to the prescriptions of UDT. The game works as follows: each player must write down an integer between $0$ and $10$. If both numbers sum to $10$ or less, then each player is paid according to the number that they wrote down. Otherwise, they are paid nothing. For example, if one player writes down $4$ and the other $3$, then the former gets paid \$4 while the latter gets paid \$3. But if both players write down $6$, then neither player gets paid. Say the human player reasons as follows:
\begin{quote}
    I don't quite know how UDT works, but I remember hearing that it's a very powerful predictor. So if I decide to write down $9$, then it will predict this, and it will decide to write $1$. Therefore, I can write down $9$ without fear.
\end{quote}
The human writes down $9$, and UDT, predicting this, prescribes writing down $1$.

This result is uncomfortable, in that the agent with superior predictive power ``loses" to the ``dumber" agent. In this scenario, it is almost as if the human's lack of ability to predict UDT (while using correct abstract reasoning about the UDT algorithm) gives the human an ``epistemic high ground" or ``first mover advantage." It seems unsatisfactory that increased predictive power can harm an agent.

There is a larger problem facing this formalism of UDT, though: even in simple examples, the algorithm is not guaranteed to work. Consider what happens if $\Algo=\texttt{const}\ \Refuse$: Then there is a proof in $\PA$ that $\Algo \neq \Pay$, and so $\Algo=\Pay$ implies \emph{anything} (by the principle of explosion). As soon as proof-based UDT proves that an agent will not take a certain policy, it concludes that taking that policy leads to the best possible outcome (because from a contradiction, anything follows). It identifies the policy that it proved to be impossible as the best one. If the outcomes are $O := [\,3, 2, 1\,]$ corresponding in $\Env$ to the policies $\Pi := \set{\Hi,\Med,\Lo}$, and the algorithm is $\Algo := \texttt{const}\ \Lo$, then proof-based UDT may misidentify $\Med$ as the best available policy.

As discussed by \citet{Benson-Tilsen:2014}, this problem is especially interesting in the case where $\Algo=\texttt{UDT(\Quote{\Env},\Quote{\Algo})},$\footnote{While this may seem circular, such a thing is possible by quining \citep{Hofstadter:1979}, e.g.\ via Roger's fixed-point theorem \citep{Rogers:1987}.} e.g.\ when the UDT algorithm is identifying the best policy \emph{available to the UDT algorithm itself} while embedded in some environment $\Env$. In this case, UDT does in fact select the best policy for which an outcome is provable. This follows from the consistency of $\PA$. Imagine that UDT identifies a policy $\pi$ which UDT provably does not select. Then UDT returns $\pi$---but this is a contradiction. Therefore, there is no policy $\pi$ such that $\PA$ can prove $\Algo \neq \pi$, and the best policy is reliably identified \citereset\citep{Benson-Tilsen:2014}.

However, this self-referential trickery does not change the fact that UDT can misidentify $\Med$ as the best policy available to $\texttt{const}\ \Lo$. This problem raises questions about of what it means to ask what ``would happen" if $\texttt{const}\ \Lo$ ``chose" $\Hi$. How complex does a decision algorithm have to be before calling it an ``agent"? Is a rock with the word ``Low" written on it an agent? What is as an embedding of an agent? It is not at all obvious that reasoning about the logical implications $\Algo=\pi$ is the right way to formalize counterfactual reasoning.

These problems may well need to be answered in order to formalize UDT in a stochastic setting, where it maximizes expected utility instead of searching for proofs of a certain outcome. Such an algorithm would evaluate policies \emph{conditioned} on the logical fact $\Algo=\pi$, rather than searching for logical implications. Such a formalization threatens to bring back many of the difficulties encountered by EDT in \Section{edt}: how does one deal with the case where $\Algo \neq \pi$, so that $\Algo=\pi$ is a zero-probability event? As described in \Section{edt}, the obvious answers are unsatisfactory.

In order to reason about expected utility conditioned on $\Algo=\pi$, it seems necessary to develop a better understanding of how to reason about the logical effects of a contradictory statement. If one deterministic algorithm violates the laws of logic in order to output something other than what it outputs, then how does this affect other algorithms? Which laws of logic, precisely, are violated, and how does this violation affect other logical statements?

It is not clear that these questions are meaningful, nor even that a satisfactory method of reasoning about these ``logical counterfactuals" exists. It is plausible that a better understanding of reasoning under logical uncertainty would shed some light on these issues, but a satisfactory theory of reasoning under logical uncertainty does not yet exist.\footnote{A \emph{logically uncertain} reasoner can know both the laws of logic and the source code of a program without knowing what the program outputs. For a discussion of active research on this topic, see \citet{Soares:2014c}.} Regardless, it seems that some deeper understanding of logical counterfactuals is necessary in order to give a satisfactory formalization of updateless decision theory.

\section{What If the Problem is Too Hard?} \label{sec:indirect}

The goal of this study is to formally describe what constitutes a good decision. While UDT makes some progress in this direction, and suggests a number of paths for future research, there is still some concern that the task is too difficult: What if no satisfactory formalization of logical counterfactuals exists? What if it becomes possible to implement practical smarter-than-human systems before decision theory is fully understood?

As discussed in \Section{susm}, decision procedures are not necessarily stable under reflection, and self-modifying agents approximating an unsatisfactory variant of decision theory do not necessarily converge on good decision-making behavior. If a satisfactory formalization of decision theory cannot be found, some alternative approach must be used to ensure desirable behavior. It seems possible to use some sort of ``indirectly normative" approach \mkbibparens{as per \citet[\bibstring{chapter} 13]{Bostrom:2014}} in which the intelligent agent itself is made to do the abstract work of discovering a satisfactory decision theory, by figuring out what humans would have wanted if given more time (rather than by using its own judgement).

Though an indirect approach would not require a full formalization of decision theory, it may well still require an improved understanding: if one wants to delegate the task of deciding which decision theory they would have wanted, they must at least trust the system enough to make \emph{that} decision, first. It is not clear that a modern understanding of decision theory is enough for even this.

Even if one decided that causal counterfactual reasoning is good enough, they still face the problem that CDT is not fully specified. How is the agent's action node identified in the environment? How can CDT be made to handle the fact that the agent's action node is not atomic, but made of transistors which may break or overheat? How can the set of available policies be identified? What counts as an ``observation" with respect to policy selection? If the goal were to implement a practical agent approximating CDT and then use some indirect method to have it find a better decision theory to approximate, then the focus of research may shift, but a further study of decision theory would still be required.

\section{Conclusion} \label{sec:discussion}

The goal of answering all these questions is not to identify practical algorithms, directly. Rather, the goal is to ensure that the problem of decision-making is well understood: without a formal description of what is meant by ``good decision," it is very difficult to justify high confidence in a practical heuristic that is intended to make good decisions.

A well-posed question often frames its answer: a formal description of how to identify the ``best available policy" with respect to some set of preferences in an arbitrary environment would fully characterize an ideal decision-making procedure. Idealizations are impractical, but before attempting to design a heuristic that solves a problem, it is useful to understand the solution which the heuristic is intended to approximate. \mkbibparens{For further discussion, see \citet{Soares:2014}.}

Developing an idealized understanding of decision theory may seem an insurmountable task, but the insights discussed in this paper, regarding policy selection and logical counterfactuals, give some reason for optimism. Updateless decision theory provides a new take on decision theory that addresses shortcomings of both evidential and causal reasoning. While it does not fully solve the problem of decision-making, it does point towards some promising directions.

A number of open problems remain, and many of them are concrete and approachable. We are optimistic that further decision theory research could lead to significant progress toward understanding the problem of decision-making. We remain hopeful that a sufficient understanding of the problem can be attained before the need for practical solutions arises.

\printbibliography
\end{document}